\pdfoutput=1

\documentclass[11pt]{article}

\usepackage[final]{coling}

\usepackage{times}
\usepackage{latexsym}

\usepackage[T1]{fontenc}

\usepackage[utf8]{inputenc}

\usepackage{microtype}

\usepackage{inconsolata}

\makeatletter
\ifcoling@finalcopy
 \usepackage[disable]{todonotes}
\else
 \usepackage{todonotes}
\fi
\makeatother

\usepackage{times}
\usepackage{latexsym}
\usepackage{adjustbox}
\usepackage{multirow}
\usepackage{placeins}
\usepackage{graphicx} %
\usepackage{hyperref}
\usepackage{supertabular}
\usepackage{booktabs}
\usepackage{longtable}
\usepackage{booktabs}
\usepackage{subcaption}
\usepackage{colortbl}
\usepackage{cleveref}
\usepackage[framemethod=tikz]{mdframed}

\definecolor{mygray}{gray}{0.9}

\newcommand\codesnippet[2][4.5cm]{%
  \begin{mdframed}[roundcorner=7pt, backgroundcolor=mygray, fontcolor=purple, linewidth=0pt]
  \parbox[t][#1][t]{\linewidth}{#2}
  \end{mdframed}%
}

\newcommand{\mc}{\textit{multiple choice}}
\newcommand{\tf}{\textit{true - false}}
\newcommand{\num}{\textit{number}}
\newcommand{\ftg}{\textit{fill the gap}}
\newcommand{\bin}{\textit{binary}}
\newcommand{\oq}{\textit{open question}}
\newcommand{\other}{\textit{other}}
\newcommand{\ITA}{Invalsi ITA}
\newcommand{\MATE}{Invalsi MATE}
\newcommand{\OMATE}{Olimpiadi MATE}

\newcommand{\llamantino}{\textit{llamantino 2 70b chat}}
\newcommand{\llamadue}{\textit{llama 2 70b chat}}
\newcommand{\camoscio}{\textit{camoscio 2 70b instruct}}
\newcommand{\llamatreotto}{\textit{llama 3 8b instruct}}
\newcommand{\llamatreunootto}{\textit{llama 3.1 8b instruct}}
\newcommand{\llamatresett}{\textit{llama 3 70b instruct}}
\newcommand{\llamatreunosett}{\textit{llama 3.1 70b instruct}}
\newcommand{\llamatreunoquattro}{\textit{llama 3.1 405b instruct}}
\newcommand{\anita}{\textit{anita 8b dpo}}
\newcommand{\mistral}{\textit{mistral instruct}}
\newcommand{\mixtral}{\textit{mixtral instruct}}
\newcommand{\minerva}{\textit{minerva 3b}}
\newcommand{\llamaduesettbase}{\textit{llama 2 70b}}

\title{The Invalsi Benchmarks: measuring the Linguistic and Mathematical understanding of Large Language Models in Italian}
\author{
  Giovanni Puccetti\\
  ISTI CNR\\\And
  Maria Cassese\\
  ISTI CNR\\
  \texttt{\{giovanni.puccetti,maria.cassese,andrea.esuli\}@isti.cnr.it} \\\And
  Andrea Esuli\\
  ISTI CNR\\}

\begin{document}
\maketitle

\begin{abstract}
  While Italian is a high-resource language, there are few Italian-native benchmarks to evaluate generative Large Language Models (LLMs) in this language. This work presents three new benchmarks: \MATE{} to evaluate models performance on mathematical understanding in Italian, \ITA{} to evaluate language understanding in Italian and \OMATE{} for more complex mathematical understanding. 

  The first two benchmarks are based on the Invalsi tests, which are administered to students of age between 6 and 18 within the Italian school system and have been validated by several experts in teaching and pedagogy, the third one comes from the Italian high school math Olympics.

  We evaluate 10 powerful language models on these benchmarks and find that they are bound by 71\% accuracy on \MATE{}, achieved by Llama 3.1 70b instruct and by 88\% on \ITA{}. For both \MATE{} and \ITA{} we compare LLMs with the average performance of Italian students to show that Llama 3.1 is the only one to outperform them on \MATE{} while most models do so on \ITA{}, we then show that \OMATE{} is more challenging than \MATE{} and the highest accuracy, achieved by Llama 3.1 405b instruct accuracy is 45\%.

  We will make data and evaluation code openly available upon acceptance of the paper.
\end{abstract}

\section{Introduction}

The evaluation of Large Language Models (LLMs) is a complex task due to the general purpose nature of these systems \citep{gehrmann2023bench}. 
Evaluating different abilities requires both different benchmark datasets and evaluation metrics and references. There is therefore need for multifaceted evaluation harnesses to perform all-round evaluation of these models.

In this work we propose three benchmark datasets meant to evaluate language models on mathematical knowledge and language understanding in Italian: \MATE{}, \ITA{}, and \OMATE{}. The first two are based on the Invalsi tests, public tests that are used to assess students' skills from middle school to high school in the Italian school system, the former is meant to assess a language model's ability to perform math reasoning and the second to assess its language understanding ability. The third one, \OMATE{}, based on the Italian national math Olympics, is meant to extend \MATE{} providing more difficult questions.

\begin{figure}[t]
    \includegraphics[width=\linewidth]{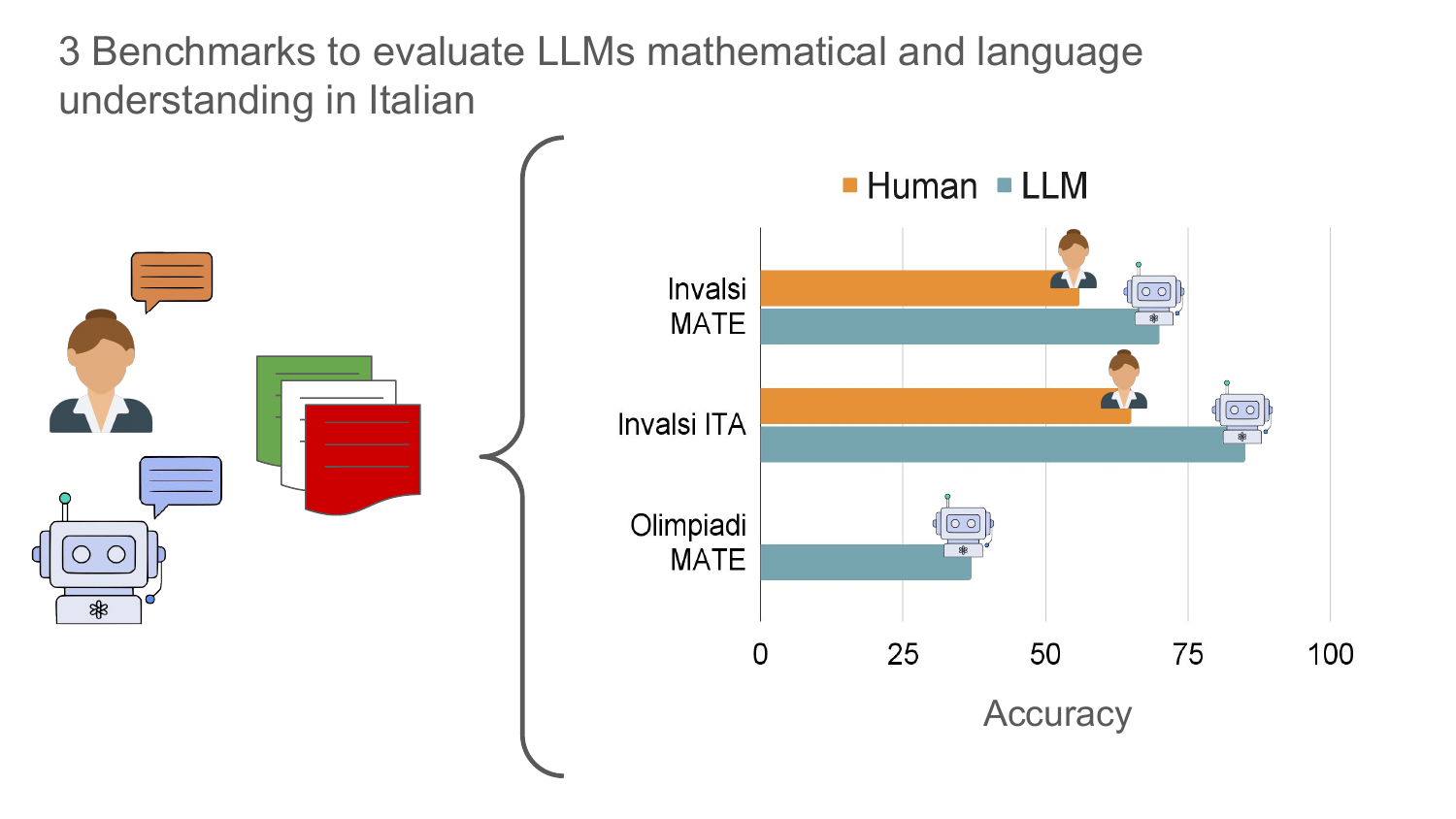}
    \caption{We show that LLMs perform better than human students on Mathematical and Language understanding in Italian.}
    \label{fig:graphical_abstract}
\end{figure}

For the questions based on the Invalsi tests, we also compare the performance of language models with that of students of different ages across Italy, and we find that language models outperform them in Italian as well, as shown in \Cref{fig:graphical_abstract}.

The contribution of this work is threefold:
\begin{enumerate}
    \item \textbf{three benchmarks for Math and Language Understanding in Italian} (two for Math, one for Language) that are the first natively Italian benchmarks of this kind;\footnote{There are benchmarks on translated datasets, however, these are not fully documented or openly available \citep{jiang2024mixtral}.}
    \item \textbf{the evaluation of 10 powerful LLMs on these benchmarks}, including Llama 3.1 405b instruct.
\end{enumerate}

We evaluate 4 kinds of models:\footnote{We write pre-trained on English to indicate models with pre-training data mostly in English, this does not exclude the presence of non-English data, intentional or not, e.g. Llama includes Wikipedia in many languages \citep{touvron_llama_2023}.}

\begin{figure}[t]
    \centering
    \includegraphics[width=\linewidth]{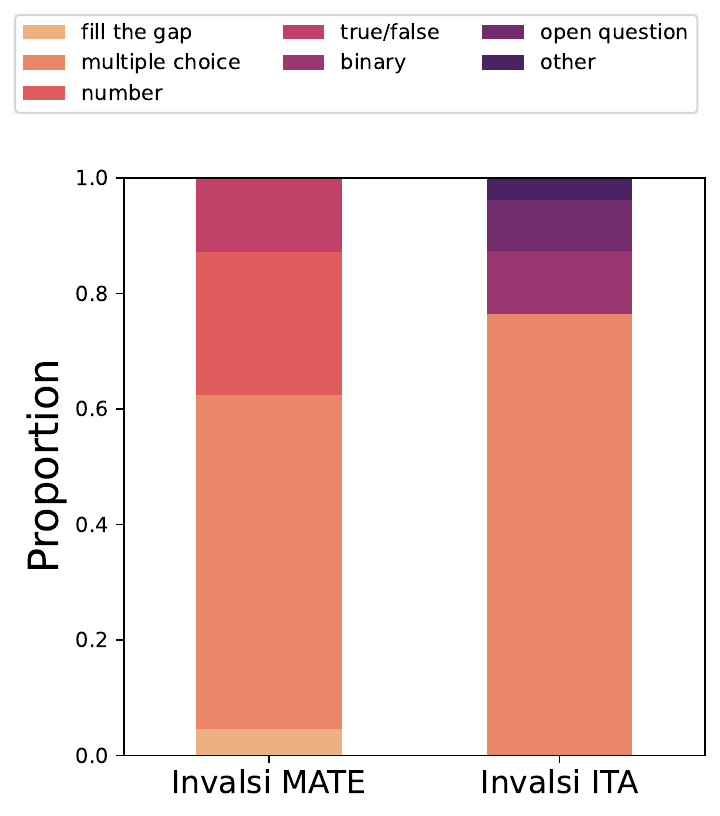}
    \caption{The distribution of Question types in \MATE{} and \ITA{}.}
    \label{fig:invalsi_type_dist}
\end{figure}

\begin{figure}[t]
    \centering
    \includegraphics[width=\linewidth]{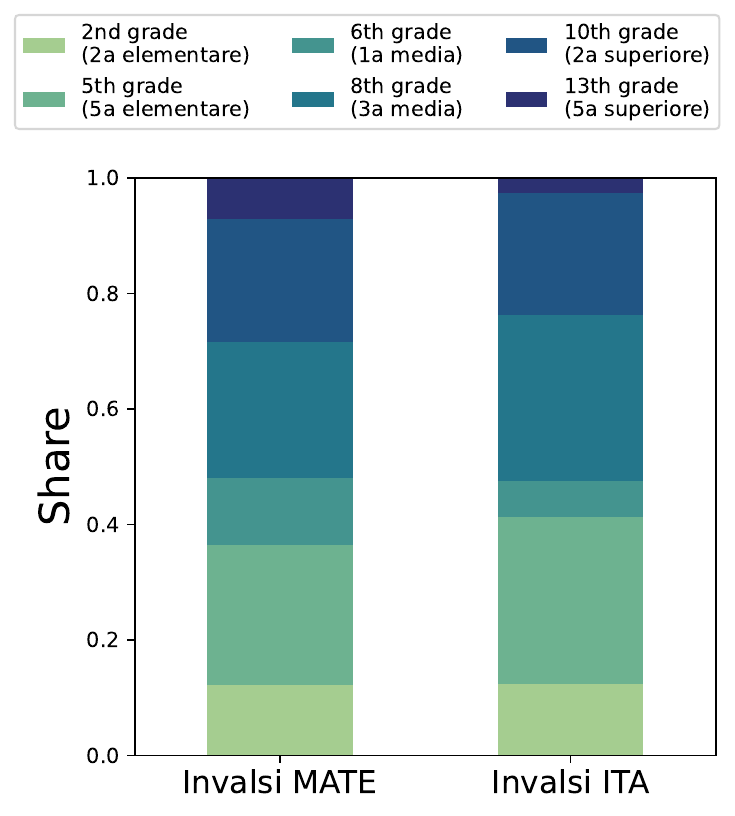}
    \caption{The distribution of Question types in \MATE{} and \ITA{}.}
    \label{fig:invalsi_grade_dist}
\end{figure}

English pre-trained and English fine-tuned, English pre-trained and Italian fine-tuned, Multilingual pre-trained and Multilingual fine-tuned, Italian pre-trained.\footnote{This category is currently only composed of one model, Minerva 3b, still relatively small and with limited performance.}

Moreover, we contribute a fine-tuning of LLama 2 70b on the Camoscio dataset \citep{santilliCamoscioItalianInstructiontuned2023}, this provides us with a strong model for the case of models pre-trained in English and fine-tuned in Italian, however, we find that fine-tuning is not sufficient to overcome the gap created by a multilingual pre-training.

Indeed, models with mixed language pre-training that on English-based evaluations are comparable to models with English-only pre-training, do perform better by a non-negligible margin on Italian-based evaluations. Nevertheless, we find that larger models trained on more text, e.g. Llama 3.1, outperform multilingual models.

The rest of the document is organized as follows \Cref{sec:related_work} presents the related work and \Cref{sec:dataset_description} describes the benchmarks. Evaluation paradigm along with the models we evaluate are described in \Cref{sec:evaluation} and the results in \Cref{sec:results_invalsi_math}, \Cref{sec:results_invalsi_ita}, and \Cref{sec:results_olimpiadi_math}. Finally, we compare the models' performance with the results of Italian students in \Cref{sec:students_compared} and draw conclusions in \Cref{sec:conclusions}.

\section{Related work}
\label{sec:related_work}

The development of open, English-first, Large Language Models is improving with several recent open weights releases \citep{touvron_llama_2023, llama3modelcard} and also fully open source ones \citep{bidermanPythiaSuiteAnalyzing2023, groeneveldOLMoAcceleratingScience2024, black-etal-2022-gpt}.

There are also open weights releases of multilingual language models, focused on languages spoken within the European Union (among other continents), French, Spanish, Portuguese, etc., such as Mistral \citep{jiang2023mistral} and Mixtral \citep{jiang2024mixtral}, as well as in other non-English languages such as Arabic \citep{sengupta2023jais} and Chinese \citep{ai2024yi}; we refer to \citep{llm-survey} for a more extensive review.\footnote{There are also models meant to work over hundreds of low resources languages \citep{ustun-etal-2024-aya}, however, that is beyond the scope of this work.}

There are no examples of Language models with 7 billion or more parameters pre-trained in Italian, however initial efforts towards such models are starting, most prominently Minerva, the only previous experiment is IT5 \citep{sarti-nissim-2022-it5} which, while it notably came earlier, it is older and smaller compared to current LLMs.

Nevertheless, several examples of fine-tunes on Italian are available. DanteLLM \citep{bacciu-etal-2024-dantellm-lets} is a chat fine-tune version of Mistral and an extension of Fauno \citep{bacciu_fauno_2023}. LLamantino is an example of continued pre-training and fine-tuning of LLaMA 2 models on Italian \citep{basile2023llamantino} using QLora \citep{dettmers2023qlora} the same developers later trained Anita \citep{polignano2024advanced} based on Llama 3 \citep{llama3modelcard}. Camoscio \citep{santilliCamoscioItalianInstructiontuned2023} is an Italian replica of Alpaca and ExtremITA is a fine-tune dedicated to the 2023 EVALITA challenge \citep{hromei2023extremita}.

The Occiglot family of models\footnote{\href{https://occiglot.eu/}{https://occiglot.eu/}} focuses on several languages spoken within Europe, including Italian. 
Further examples are, Cerbero \citep{galatolo_cerbero-7b_2023} and Maestrale \footnote{\href{https://huggingface.co/mii-llm/maestrale-chat-v0.4-beta}{https://huggingface.co/mii-llm/}}. With the exception of Llamantino, which also releases models with 13 and 70 billion parameters, all other fine-tunes are models with at most 7 Billion parameters using quantization \citep{int8dettmers} and Lora for fine-tuning \citep{hu2022lora}.

\subsection{Benchmarks}

Along with a few language models pre-trained on Italian, few benchmarks are explicitly thought for evaluating LLMs in this language, and in several cases, they are obtained by translating existing benchmarks for English instead of developing new ones.

To benchmark IT5 \citet{sarti-nissim-2022-it5} collect a dataset of Italian news for a summarization task, newssum-it, while other benchmarks are obtained by translating existing English datasets, as is done for squad-it \citep{squad_it}.

The most relevant exception is UINAUIL, a set of benchmarks to evaluate language understanding in Italian \citep{basile-etal-2023-uinauil}, based on the tasks presented at several EVALITA campaigns.

There are also a few multilingual benchmarks that include Italian \citep{hardalov-etal-2020-exams,das2024examsv}, but, to the best of our knowledge, neither mathematical nor linguistic understanding is included.

\section{Benchmark Description}
\label{sec:dataset_description}

The Invalsi tests are country-wide assessments designed to monitor the average performance of students over the years, administered multiple times from ground school through high school. The results of these tests have been used in several population studies \citep{invalsi1, invalsi2, invalsi3}, however, to the best of our knowledge, their use to benchmark Language Models' performance in Italian is unprecedented, there is work that was preprint after our own \cite{mercorio2024disceautdeficereevaluating}, and shows consistent results on a much smaller dataset.

\begin{table*}[ht]
\centering
\adjustbox{max width=\linewidth}{
\begin{tabular}{llcccc}
\toprule
Question Type & & \textit{ALL} & \mc{} & \tf{} & \num{} \\
N. Questions & & 400 & 244 & 54 & 102 \\
\midrule
Base Model & Model & \multicolumn{4}{c}{Accuracy} \\
\midrule
llama 3.1 70b &  \llamatreunosett{}    & \textbf{0.71} (± 0.01) & \textbf{0.70} (± 0.01) & \textbf{0.64} (± 0.04) & \textbf{0.78} (± 0.04) \\
\midrule
\multirow{2}{*}{mistral} & \mixtral{} & \underline{0.55} (± 0.02) & \underline{0.49} (± 0.03) & \underline{0.63} (± 0.07) & \underline{0.66} (±  0.04)\\
& \mistral{} & 0.44 (± 0.03) & 0.34 (± 0.03) & 0.59 (± 0.07) & 0.63 (± 0.07) \\
\midrule
\multirow{3}{*}{llama 2 70b} & \camoscio{} &  0.49 (± 0.02) & 0.43 (± 0.03) & 0.59 (± 0.07) & 0.62 (± 0.05)\\
& \llamantino{} & 0.47 (± 0.02)& 0.41 (± 0.03) & 0.54 (± 0.07) & 0.61 (± 0.05) \\
& \llamadue{} & 0.43 (± 0.02)& 0.40 (± 0.03) & 0.48 (± 0.07) & 0.52 (± 0.05) \\
\midrule
llama 3 8b & \anita{} & 0.47 (± 0.02)& 0.40 (± 0.03) & 0.61 (± 0.07) & 0.55 (± 0.05) \\
llama 3.1 8b &  \llamatreunootto{}  & 0.5 (± 0.01) & 0.45 (± 0.01) & 0.59 (± 0.04) & 0.59 (± 0.07)\\
\midrule
minerva & \minerva{} 3b & 0.20 (± 0.04) & 0.22 (± 0.04) & 0.50 (± 0.07) & 0.32 (± 0.05) \\
\midrule
- & \textit{random} & 0.28 & 0.25 & 0.5 & 0.25 \\
\bottomrule
\end{tabular}}
\caption{Models 0-Shot accuracy on Invalsi MATE, likelihood based evaluation. In \textbf{bold} the highest accuracy in each column and \underline{underlined} the second highest.}
\label{tab:invalsi_mate_likelihood}
\end{table*}

These tests are of three types: Mathematical Understanding, Language Understanding, and English Understanding. Given our current focus on Italian, we discard the last one.

The National Math Olympics consists of tests designed for students between the ages of 14 and 18. These tests are more challenging than the Invalsi tests and are generally only administered to students willing to test themselves. The questions are more complex in several ways; they require more reasoning, are often more open-ended, and tend to involve more advanced topics.

\textbf{To create each benchmark we have collected the data from their original sources and we gathered the full history of questions and answers.} Two annotators\footnote{One of the annotators holds an MSc in Mathematics and the other holds an MSc in Computer science, this gives them sufficient knowledge to evaluate both \MATE{} and \OMATE{} while for \ITA{} the questions are addressable by any native speaker of Italian that went through the mandatory education.} have manually checked all the samples to control if there were mistakes and if the collected answers were right.

In all the tests, a small share of the questions also have a visual component. In this work, we focus on textual-only questions and we exclude them. We plan to develop a follow-up benchmark for Italian Visual Language Models (VLMs).

Looking ahead, the annual cycle of Invalsi and Math Olympics allows for a periodic update of the benchmarks based on new test releases. 
This would provide a continuous stream of test sets resilient to data leakage and allow for an increase in the benchmarks' size.

\subsection{\MATE{}}

The math test based on Invalsi is composed of several questions that fall into four types:
\begin{itemize}
  \item \mc{}: the student is asked to pick the correct answer among four candidate answers;
  \item \tf{}: the student is asked to assess whether a given statement is True or False;
  \item \num{}: the student is asked a question that admits a given number as an answer;
  \item \ftg{}: the student is asked to fill one or more missing words in a given text, based on logical and mathematical reasoning.
\end{itemize}

See \Cref{app:question_examples} for references about the questions for examples of each question type.

To evaluate LLMs performance on these tasks we use a likelihood-based approach. We compare the likelihood of each possible completion and select the highest one as the answer chosen by the LLM. While only \mc{} and \tf{} questions are naturally meant to be evaluated in this way, we recast the \num{} questions by manually adding wrong options next to the correct answers.

We exclude the \ftg{} questions from this kind of evaluation since they are hard to adapt to this setting. The \MATE{} benchmark is composed of a total of 420 questions. \Cref{fig:invalsi_type_dist} shows the share of questions of each type. The \mc{} questions make up about 58\% of the questions, the second most numerous are \num{} questions then \tf{} while there are only 20 \ftg{} questions.

\begin{table*}[t]
\centering
\adjustbox{max width=\linewidth}{
\begin{tabular}{llccc}
    \toprule
    Question Type & & \textit{ALL} & \mc{} & \bin{} \\
    N. Questions & & 1117 & 977 & 140 \\
    \midrule
    Base Model & Model & \multicolumn{3}{c}{Accuracy} \\
    \midrule
    llama 3.1 70b &  \llamatreunosett{}    & \textbf{0.88} (± 0.01) & \textbf{0.9} (± 0.01) & \textbf{0.75} (± 0.04) \\

    \midrule
    \multirow{2}{*}{mistral} & \mixtral{}        & \underline{0.80} (± 0.01) & \underline{0.82} (± 0.01) &  \underline{0.69} (± 0.04) \\
     & \mistral{}   & 0.49 (± 0.01) & 0.60 (± 0.02) &  0.51 (± 0.04) \\
    \midrule
    \multirow{3}{*}{llama 2 70b} & \camoscio{} & 0.78 (± 0.01) & 0.78 (± 0.01) &  0.67 (± 0.04) \\
    & \llamantino{}   & 0.74 (± 0.01) & 0.75 (± 0.01) &  0.63 (± 0.04) \\
    & \llamadue{} & 0.72 (± 0.01) & 0.73 (± 0.01) &  0.64 (± 0.04) \\
    \midrule
    llama 3 8b & \anita{} & 0.71 (± 0.01) & 0.72 (± 0.01) &  0.66 (± 0.04) \\
    llama 3.1 8b & \llamatreunootto{} & 0.71 (± 0.01) & 0.72 (± 0.01) & 0.6 (± 0.04)\\
    \midrule
    minerva & \minerva{} & 0.30 (± 0.01) & 0.25 (± 0.01) &  0.54 (± 0.04) \\
    \midrule
    - & \textit{random} & 0.27 & 0.25 & 0.44 \\ 
    \bottomrule
    \end{tabular}}
\caption{Models 0-Shot accuracy on Invalsi ITA, likelihood based evaluation. In \textbf{bold} the highest accuracy in each column and \underline{underlined} the second highest.}
\label{tab:invalsi_ita_likelihood}
\end{table*}

\subsection{\ITA{}}

Language understanding tests are based on a piece of text that the student reads. 
The text can be as simple as a kid novel for younger students, or as complex as an essay or a journal article for older students. The student is then presented with a set of questions concerning the passage.

Similarly to the math section, we group the questions for Italian Language Understanding in four classes:
\begin{itemize}
  \item \mc{}: the student is asked to pick the correct answer among four candidate answers;
  \item \bin{}: the student is asked to assess a binary property of a statement, e.g. True - False, Before - After, etc.
  \item \oq{}: the student is asked to identify a passage in the text that answers the question;
  \item \other{}: A small share of questions belong to open-ended questions with varying scopes that are hard to put under a single label.
\end{itemize}

See \Cref{app:question_examples} for references about the questions and for examples of each question type. The \ITA{} benchmark is composed of a total of 1264 questions. \Cref{fig:invalsi_type_dist} shows the share of questions of each type. The \mc{} questions make up more than 76\% of the questions, the second in quantity are \bin{} questions then \oq{}, and finally, there are only 44 \other{} questions.

One of the differences between \MATE{} and \ITA{} is that the questions in the latter set often concern a longer text passage that needs to be processed entirely by the model as context to answer the questions. This makes the questions more computationally demanding and changes the type of actions needed to answer them. While \MATE{} questions require reasoning to be answered, \ITA{} questions require the ability to retrieve information from longer texts and leverage general knowledge.

\subsection{\OMATE{}}

The \OMATE{} benchmark contains only \mc{} questions with 5 possible choices A, B, C, D, and E. While this might appear to simplify the dataset, these questions are inherently more difficult, as also confirmed in our evaluation. The benchmark consists of 619 questions. %

\subsection{Distribution by Grade}

The Invalsi tests are taken by students of varying ages between first grade, 6 years old, and 13th grade, 18 years old, the distribution of questions by grade is shown in \Cref{fig:invalsi_grade_dist}. The questions are fairly evenly distributed across all grades in \MATE{} while for \ITA{} grades 6th and 13th  are less present, reflecting the distribution of data in original tests. In \Cref{sec:results_invalsi_math} and \Cref{sec:results_invalsi_ita} we show how models' performance varies across different grade levels.

The Math Olympics also have different tests based on the students' age. In particular, 9th and 10th-grade students have one set of questions, while 11th, 12th, and 13th-grade students have another. We refer to these sets as B and T respectively. Set B is composed of 267 questions, and set T is composed of 352 questions. Together they form the \OMATE{} benchmark.

\section{Evaluation}
\label{sec:evaluation}

\begin{figure*}[ht]
    \centering
    \begin{subfigure}{0.8\textwidth}
        \includegraphics[width=\linewidth]{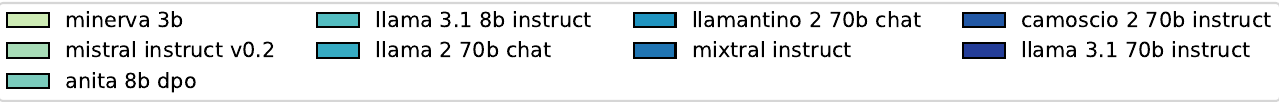}
    \end{subfigure}
    \begin{subfigure}{0.47\textwidth}
        \includegraphics[width=\linewidth]{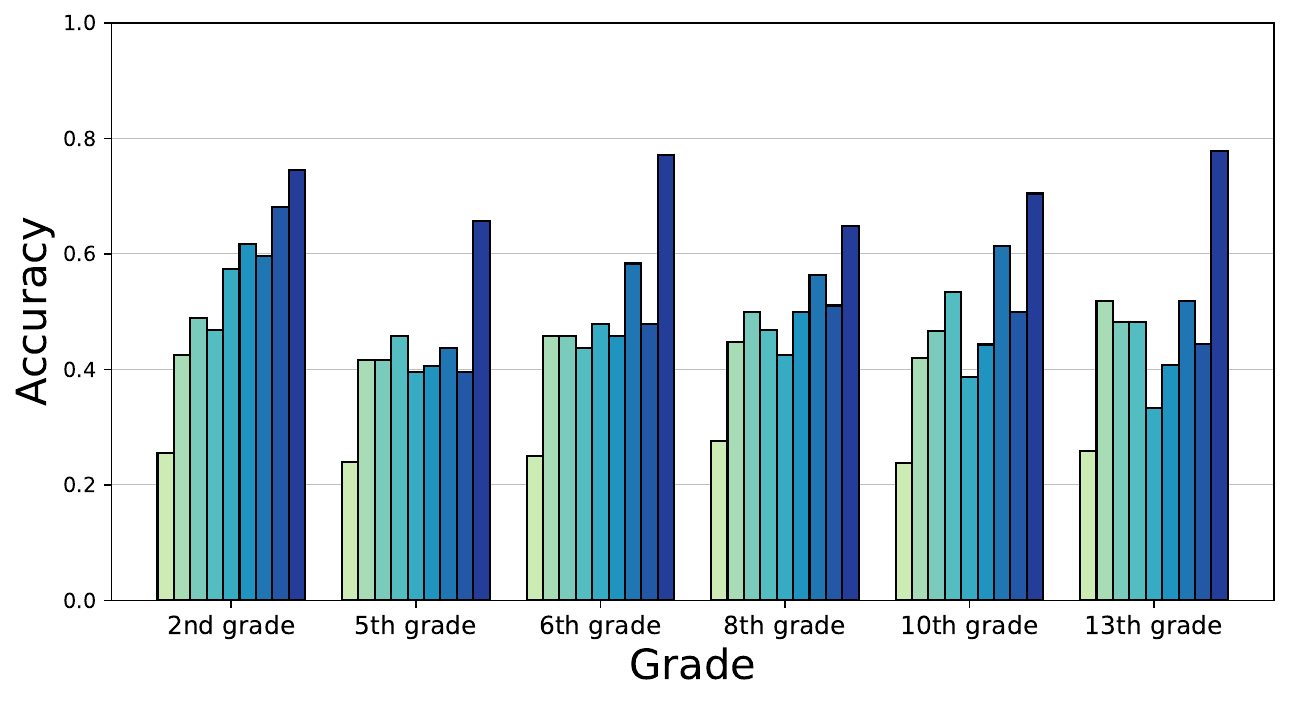}
        \caption{Invalsi MATE}
        \label{subfig:invalsi_mate_perf_grade_dist}
    \end{subfigure}
    \begin{subfigure}{0.47\textwidth}
        \includegraphics[width=\linewidth]{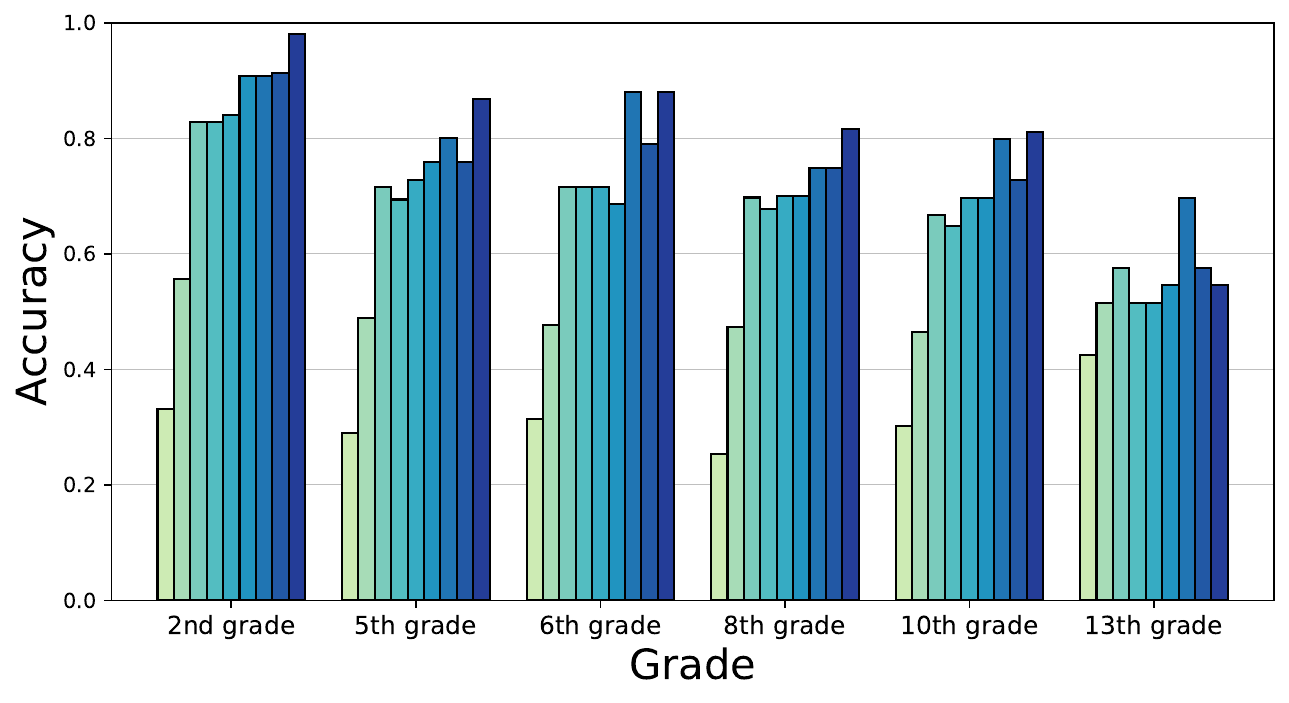}
        \caption{Invalsi ITA}
        \label{subfig:invalsi_ita_perf_grade_dist}
    \end{subfigure}

    \caption{The performance stratified for different grades, in (\subref{subfig:invalsi_mate_perf_grade_dist}) for Invalsi MATE and in (\subref{subfig:invalsi_ita_perf_grade_dist}) for Invalsi ITA.}
    \label{fig:invalsi_perf_grade_dist}
\end{figure*}

\subsection{Models}
\label{sec:models}

We divide the models we evaluate into 4 categories:\footnote{In \emph{italics} the names we use for each of them in tables, see also \Cref{app:model_naming} for more details.}

\paragraph{English pre-trained and Italian fine-tuned} These models are pre-trained on English and then fine-tuned on Italian. We evaluate our own fine-tune of Llama 2 70b on the Camoscio dataset \citep{santilliCamoscioItalianInstructiontuned2023}, and we name it Camoscio 2 70b (\camoscio{}). 
We also evaluate LLamantino 70B chat (\llamantino{}) which is a fine-tune of LLama 2 70B chat on the ultrachat dataset translated in Italian \citep{basile2023llamantino}. And the smaller Anita (\anita{}) is a model built upon Llama 3 8b.

\paragraph{Multilingual pre-trained and Multilingual fine-tuned} These models are both pre-trained and fine-tuned on multilingual datasets.
We consider the two most effective models of this kind that have been trained on Italian and languages close to it, Mistral-Instruct-v0.2 (\mistral{}) \citep{jiang2023mistral} and Mixtral-Instruct-v0.1 (\mixtral{}) \citep{jiang2024mixtral}.

\paragraph{English pre-trained and English fine-tuned} The developers of these models have intentionally removed all non-English texts in their pre-training corpus, except for selected sources, e.g. Wikipedia the models of this kind we check are those from the Llama 2 and Llama 3.1 family \citep{touvron_llama_2023} Llama 2 70b chat (\llamadue{}), Llama 3.1 8b instruct (\llamatreunootto{}), Llama 3.1 70b instruct (\llamatreunosett{}) and Llama 3.1 405b instruct (\llamatreunoquattro{}).

\paragraph{Italian pre-trained} There is currently only one model pre-trained only on English and Italian, with a focus on Italian, Minerva 3b (\minerva{}), however, its largest version encompasses 3 billion parameters and thus it compares poorly to the other larger models we test.\footnote{We test it nevertheless because in the future, larger models of this same family will be available and it will be a natural step forward to test them on our benchmark.}

\subsection{Results on Invalsi MATE}
\label{sec:results_invalsi_math}

\begin{table*}[ht]
    \centering
    \adjustbox{max width=\linewidth}{
    \begin{tabular}{llccc}
    \toprule
    Dataset Split & & \textit{ALL} & B & T \\
    N. Questions & & 619 & 267 & 352 \\
    \midrule
    Base Model & Model & \multicolumn{3}{c}{Accuracy} \\
    \midrule
    llama 3.1 405b & \llamatreunoquattro{} & \textbf{0.45} (± 0.02) & \textbf{0.46} (± 0.03) & \textbf{0.45} (± 0.03)\\
    llama 3.1 70b & \llamatreunosett{} & \underline{0.34} (± 0.02) & \underline{0.34} (± 0.03) & \underline{0.34} (± 0.03)\\
    \midrule
    \multirow{2}{*}{mistral} & \mixtral{} & 0.26 (± 0.02) & 0.24 (± 0.02) & 0.28 (± 0.02) \\
    & \mistral{} & 0.27 (± 0.02) & 0.29 (± 0.03) & 0.26 (± 0.02) \\
    \midrule
    \multirow{2}{*}{llama 2 70b} & \camoscio{} & 0.22 (± 0.02) & 0.23 (± 0.03) & 0.21 (± 0.02)\\
    & \llamantino{} & 0.25 (± 0.02) & 0.23 (± 0.03) & 0.26 (± 0.02) \\
    \midrule
    llama 3 8b & \anita{} & 0.23 (± 0.02) & 0.26 (± 0.03) & 0.22 (± 0.02)\\
    llama 3.1 8b & \llamatreunootto{} & 0.23 (± 0.02) & 0.2 (± 0.02) & 0.24 (± 0.02)\\
    \midrule
     & \textit{random} & 0.2 & 0.2 & 0.2 \\
    \bottomrule
    \end{tabular}}
    \caption{Models 0-Shot accuracy on Olimpiadi MATE, likelihood based evaluation. In \textbf{bold} the highest accuracy in each column and \underline{underlined} the second highest.}
    \label{tab:olimpiadi_mate_likelihood}
    \end{table*}

\paragraph{Likelihood Based Evaluation}

We measure models' performance on our dataset using a likelihood-based evaluation method, as done in \cite{cobbe2021training, eval-harness}. See \Cref{app:prompts} for details about the prompt we use, more details will also be available in the code we will release upon publication.

\Cref{tab:invalsi_mate_likelihood} shows the results of this evaluation, each column represents the accuracy on a specific type of question, and the \textit{ALL} column the accuracy over the whole set.

The most clear result is how \llamatreunoquattro{} outperforms every other model by a large margin, the results in bold in \Cref{tab:invalsi_mate_likelihood} show how its accuracy is more than 10\% higher than all others across all tasks. However, the Llama 3.1 model family is trained on 15 trillion tokens, reportedly way more than all other models we consider. Looking at the group of the other models trained with comparable training resources, there are two main takeaways:

\begin{itemize}
  \item \textbf{language models pre-trained mostly in English, perform worse than smaller models pre-trained on multilingual data:} \mistral{} performs better than comparably large models on \MATE{} and \mixtral{} outperforms larger models;
  \item \textbf{further training on Italian does not mitigate this performance gap}: our own \camoscio{} and \llamantino{} only show marginal gains on mathematical understanding when compared to \llamadue{} and the same holds for \anita{} when compared to \llamatreotto{} showing that fine-tuning on Italian does not appear to help on mathematical understanding.
\end{itemize}

Indeed, \mixtral{} performs better than all other models, including all those based on \llamaduesettbase{} which are larger in terms of parameters, and all those that had extra training on Italian, \camoscio{} and \llamantino{}.

\paragraph{Results Break-Down by Grade}

The Invalsi dataset is stratified by students' grade allowing us to look at model performance on each separate grade, \Cref{subfig:invalsi_mate_perf_grade_dist} shows the performance when only answering questions of a given grade. The largest difference between two grades for a single model is below 10\%. However, there is a clear difference between 1st and 2nd grade where models perform better, and the remaining ones: 5th, 6th, 8th, 11th, and 13th.

\subsection{Results on Invalsi ITA}
\label{sec:results_invalsi_ita}

We measure the same LLMs also on \ITA{}, \Cref{tab:invalsi_ita_likelihood} shows their performance on each question type as well as for all of them together.

Similar to what happens for \MATE{} \llamatreunosett{} is the strongest model, however, the performance gap is sensibly lower in this case. While there is a 15\% gap between the accuracy of the second best \mixtral{} on \MATE{} for \ITA{} this gap is about 8\%. This suggests that \textbf{when working on Italian, LLMs can retain reasoning skills acquired in English but suffer a drop in language understanding.}

Excluding \llamatreunosett{}, similar results also apply to \MATE{}: multilingual pre-training provides strong performance improvements in Italian, and similar improvements are hard to achieve with extra training in this language.

\paragraph{Results Break-Down by Grade}

\ITA{} also provides the grade of the students who answered a given question, therefore we can study performance by grade.
\Cref{subfig:invalsi_ita_perf_grade_dist} shows the accuracy of the models on the questions of each grade, unlike for \MATE{} we can see a clear descending pattern: LLMs find answering to later grades questions harder and this happens uniformly for all the models we test.

\begin{table*}[t]
\centering
\adjustbox{max width=\linewidth}{
\begin{tabular}{llcccccccc}\toprule
Subject & Question type &\multicolumn{5}{c}{Grade}  & All grades & N. Questions \\\midrule
 & &2 &5 &6 &8 &10 & \\
 \midrule
 & \mc{} &0.55 &0.54 &0.48 &0.58 &0.44 &0.53 & 831 \\
Mathematics & \tf{} &0.53 &0.68 &0.51 &0.67 &0.62 &0.65 & 325 \\
 & \textit{multi + t - f}&0.55 &0.58 &0.48 &0.62 &0.51 &0.56 & 1156 \\
  \midrule
 & \mc{} &0.56 &0.63 &0.57 &0.68 &0.63 &0.62 & 1004 \\
Italian & \bin{} &0.61 &0.69 &0.73 &0.74 &0.72 &0.70 & 611 \\
 & \textit{multi + bin} &0.57 &0.65 &0.61 &0.70 &0.67 &0.65 & 1615 \\
\bottomrule
\end{tabular}}
\caption{Students' accuracy on Invalsi tests, \mc{} and and \tf{} questions (together in \textit{multi + t - f}) for \MATE{} and \mc{} and \bin{} questions (together in \textit{multi + \bin{}}) for \ITA{}. Source: \href{https://www.gestinv.it/}{https://www.gestinv.it/}.}
\label{tab:students_scores}
\end{table*}

\subsection{Results on \OMATE{}}
\label{sec:results_olimpiadi_math}

\Cref{tab:olimpiadi_mate_likelihood} shows the evaluation of models on the \OMATE{} benchmark, unlike \MATE{} we see that models' accuracy is below 45\%, which is only reached by \llamatreunoquattro{}, this model is 5 to 50 times larger than all other ones and very costly to use.\footnote{To run \llamatreunoquattro{}, due to its size, we have to use a costly infrastructure, 4x4 64GBA100 and therefore we only run it on this benchmark, which is the one where other models struggle the most.} Looking at the breakdown by B and T, the two sets of questions meant for different age groups, we see that the performance is similar across the two sets. While there is a clear best model also for this benchmark \llamatreunoquattro{} and \llamatreunosett{} are the only two models with accuracy above 30\%.

Interestingly, \mistral{} outperforms \mixtral{} on \OMATE{}, although only marginally, unlike on the Invalsi benchmarks. This suggests that \mistral{} has undergone deeper fine-tuning for mathematical understanding and confirms that besides massive extra training, e.g. Llama 3.1, multilingual first models perform better in Italian.

We also experiment with different evaluation approaches, either based on human inspection of the output, see \Cref{app:human_eval} or on pattern matching evaluation, see \Cref{app:pattern_match_eval}.

\section{Comparison with students' results}
\label{sec:students_compared}

We have put together the aggregated evaluations of students' answers from Invalsi, collected from tens of thousands of students per grade\footnote{No data is currently publicly available for the 13th grade.}, per year \cite{invalsi2}.

We have obtained accuracy scores by question type and grade for both Italian and Mathematics tests, reported in \Cref{tab:students_scores}. 
It was not possible to exactly select only the questions that compose \ITA{} and \MATE{}, so this evaluation also accounts for questions that include images.

For this reason, we cannot make a rigorous statistical comparison, yet the values are qualitatively comparable.

Given the caveat, it is interesting to see that on \MATE{} only \llamatreunosett{} performs better than students, with \mixtral{} close to them, all other models are worse, most notably on \mc{} questions. Differently, on \ITA{} most models perform better than students on \mc{} questions and instead worse on \bin{} ones.

Along grades, the performance of students varies more, with no clear trend, differently from the trend shown by LLMs, more so on \ITA{}. This is expected as each grade is a different population and the test for each grade is tailored to that population, while we evaluate each LLM across all grades.

\section{Conclusions and future work}
\label{sec:conclusions}

In this work we introduced three benchmarks, \MATE{}, \ITA{} and \OMATE{}, to evaluate the performance of LLMs in Italian, on mathematical and language understanding, we collected a total of 1039 questions on mathematical knowledge and 1249 on language understanding. We measured how well 10 language models perform, including our own fine-tune of LLaMa 2 70B on an Italian dataset. We find that, excluding Llama 3.1 70b (which is reportedly trained on more text than all other models we consider), multilingual models are stronger than English-first ones even when having fewer parameters.

We show that models pre-trained on multiple languages are more accurate in Italian than models pre-trained in English and that fine-tuning on Italian can't fill this performance gap. However, testing \llamatreunosett{} we show that regardless of the amount of training data, model performance in Italian remains lower than in English. We provide early results that LLMs can transfer mathematical reasoning across languages better than language understanding and we find that the gap between \llamatreunosett{} in \MATE{} is larger than in \ITA{}. Through evaluations on \OMATE{} we also show that current LLMs are currently not able to consistently solve complex mathematical problems in Italian.

Finally, we report how LLMs perform compared to the population of Italian students to show that they are close to students on \MATE{} while they outperform students on \ITA{}. Future works will integrate questions about images as well as extensions of the dataset and the validation of future stronger models. Moreover, we intend to continuously update this benchmark along with the yearly Invalsi tests held in schools.

\section{Limitations}

A limitation of this work is that \ITA{} can be almost aced by the most powerful LLMs. Nevertheless, not all the LLMs have top performance on \ITA{} and the distribution of performance is indeed varied, making this benchmark appropriate for the early stage development of the Italian LLMs ecosystem, and the exploration of size and resource optimization of state-of-art models.
Moreover, since this benchmark is based on real tests, it still provides useful insights into the current performance of LLMs compared to Italian students.
\MATE{} and specially \OMATE{} show instead to still have margin on the top performing LLMs. 

In this work we tested only openly available LLMs. This is motivated by the cost of accessing closed LLMs and also the reduced scientific value of including them into the comparison, due to their lack of specification. To address the potential performance gap with open models, we test \llamatreunoquattro{} on the hardest of our benchmarks, whose performance is reportedly comparable to the best closed models \footnote{\href{https://lmarena.ai/}{https://lmarena.ai/}}.

As with most benchmarks designed for LLMs pre-trained on web-scale datasets, there is a potential risk that one or more of the tested LLMs may have seen some of the %
questions during training. This could lead such LLMs to achieve a spurious better performance. From the descriptions of the training data of the tested LLMs it seems improbable that the content of the
tests is explicitly included since they have a very limited diffusion on the web.
Moreover, to the best of our knowledge, the answers to the questions of both Invalsi and Math Olympics are accessible only after a registration and a login. 

The release of \MATE{}, \ITA{}, and \OMATE{} exposes them to the risk that future models may be trained explicitly on them. Evaluation of future LLMs should also include a test for the presence of the content of the benchmarks in the training data. On the other hand, the annual nature of these benchmarks allows for continuous updates to the dataset, making it more difficult for models to overfit.

The comparison with students is imperfect since they are evaluated on visual questions too, which are not included in our benchmark, and currently, we can't make a per-question comparison with students, nevertheless, the large amount of students involved and the fact that Invalsi tests are specifically developed to measure the varying of students' performance over the years should mitigate this issue.

\bibliography{zotero_esuli,zotero_puccetti,biblio}

\clearpage

\appendix

\clearpage
\section{Questions Examples}
\label{app:question_examples}

\begin{center}
\footnotesize
\tablefirsthead{%
		\toprule
		   \emph{Multiple Choice} & \emph{Binary} \\
		\midrule}
	\tabletail{%
		\bottomrule
	}
	\tablelasttail{\bottomrule}
	\tablecaption{Examples of \ITA{} samples.}
 \begin{supertabular}{|m{7cm}|m{7cm}|}
             {\textbf{Testo} \newline
        Il titolo dice che Polipetto ha un problema e l’inizio del
racconto spiega di che cosa si tratta.\newline
        \textbf{Domanda} \newline
        Qual è il problema di Polipetto? \newline
        A. Non gli piace più la sua casa \newline
        B. Non può più entrare in casa sua \newline
        C. La sua casa si è riempita di animali \newline
        D. La sua casa non è più ordinata come prima} &
        {\textbf{Testo} \newline 
        Grazie all’incontro con il grande Oceano qualcosa cambia in Polipetto. \newline
         \textbf{Domanda} \newline} 
         Com’è Polipetto PRIMA di parlare con Oceano e DOPO avere parlato con lui?
Metti una crocetta per ogni riga. \newline
A. Polipetto si fa coraggio: Prima/Dopo \newline
B. Polipetto è confuso: Prima/Dopo \newline
C. Polipetto pensa che gli altri siano più bravi di lui: Prima/Dopo \newline
D. Polipetto si fida della sua idea: Prima/Dopo \newline \\
    \hline
Translation & Translation \\
    \hline
{\textbf{Text}\newline The title says that Polipetto has a problem and the beginning of the story exaplains what it is.\newline
\textbf{Domanda}\newline
Which one is Polippeto's problem?
A. He doesn't like his house \newline
B. He can't enter his house \newline
C. His house is filled with animals \newline
D. His houes is not tidy anymore} &
{\textbf{Text} \newline
After talking with great Oceano something changes in Polipetto. \newline
\textbf{Question} \newline
How is Polipetto BEFORE talking to Oceano and AFTER speaking with him? Pick one for each line. \newline
A. Polipetto motivates himself: BEFORE/AFTER \newline
B. Polipetto is confused:  BEFORE/AFTER \newline
C. Polipetto believes that others are better than he is: BEFORE/AFTER \newline
D. Polipetto trusts his idea: BEFORE/AFTER \newline
} \\
    \hline
        \textit{\textbf{Open Question}} &  \textit{\textbf{Other}} \\
    \hline
        {\textbf{Testo} \newline 
        Che cosa ci dice il racconto a proposito dell’escursione? Completa la sintesi che segue, inserendo la parola appropriata in ogni spazio. \newline
        \textbf{Domanda} \newline
"Durante l’escursione alcuni ragazzi non si accorgono di quanto camminano perché cantano e scherzano. All’arrivo, dopo esattamente cinque ore di strada. ...  (1) si siede perché è affaticato e i ... (2) gli fanno male; posa vicino a sé lo zaino con la ... (3) dentro''} & 
        {\textbf{Testo} \newline
        ''Doveva assolutamente parlarne con un amico''. Nel testo non c’è scritto che cosa ha detto Polipetto a questo amico, ma dal racconto si può capire. \newline
        \textbf{Domanda} \newline
        Che cosa può avergli detto Polipetto?
        } \\
    \hline
    Translation & Translation \\
    \hline
    {\textbf{Text} \newline What does the story say about the fieldtrip? Complete the summary that follows, adding the appropriate word in every gap. \newline\textbf{Question} \newline
    "During the fieldtrip some kids loose track of how long they walked because they sing and joke. When they arrive, after walking for 5 hours, ...(1) sits down because he is tired and his ...(2) hurt; he lays his backpack next to himself with the ...(3) inside. Miss Salici finds the lake gorgeous, while the ...(4) are not impressed."} & {\textbf{Text}\newline
    "He had to talk about it with a friend". The story does not mentioned what Polipetto said to his friend, but it can be understood from the story.\newline
    \textbf{Question} \newline
    What could have Polipetto said to him?} \\

 \end{supertabular}
 \end{center}

\clearpage
\begin{center}
\footnotesize
\tablefirsthead{%
		\toprule
		   \emph{Multiple Choice} & \emph{True - False} \\
		\midrule}
	\tabletail{%
		\bottomrule
	}
	\tablelasttail{\bottomrule}
	 \tablecaption{\textbf{Examples of \MATE{} samples.}}
 \begin{supertabular}{|m{7cm}|m{7cm}|}
{\textbf{Testo} \newline
         Elisa è uscita da casa questa mattina alle ore 8:15.
Elisa è rientrata nel pomeriggio alle ore 1:15 \newline
        \textbf{Domanda} \newline
        Quanto tempo è stata fuori casa Elisa? \newline
        A. 5 ore B. 7 ore C. 9 ore} & {\textbf{Testo} \newline
            Se moltiplichi per 2 un numero naturale e dal risultato sottrai 1, ottieni sempre un numero pari.\newline
         \textbf{Domanda} \newline
            Vero o Falso?
            }
        \\
    \hline
    Translation & Translation \\
    \hline
    {\textbf{Text}\newline
    Elisa left her house in the morning at 8:15 am. Elisa came back in the afternoon at 1:15 pm\newline
    \textbf{Question}\newline
    How long was Elisa outside?
    A. 5 hours; B. 7 hours; C. 9 hours} & {\textbf{Text}\newline
    If you multiply a natural number by 2 and subtract 1 you always get an even number.\newline
    \textbf{Question}\newline
    True or False?} \\
    \hline
        \textit{\textbf{Number}} &  \textit{\textbf{Fill the gap}}\\
    \hline
        {\textbf{Testo} \newline
        Filippo dice: per trovare il numero della mia maglietta aggiungi una decina e sei unità al numero 4. \newline
        \textbf{Domanda} \newline
        Qual è il numero della maglietta di Filippo?
        } & {\textbf{Testo} \newline
        Luca lancia due dadi a sei facce non truccati. \newline
        \textbf{Domanda} \newline
        Completa la frase inserendo una delle seguenti espressioni. \newline
         - maggiore della \newline
         - minore della \newline
         - uguale alla \newline
        La probabilità che la somma dei punti sia 12 è .... probabilità che la somma sia 2.
        }\\
    \hline
    Translation & Translation \\
    \hline
    {\textbf{Text}\newline
    Filippo says: to find out the number of my t-shirt add one tens and six units to 4\newline
    \textbf{Question}\newline
    What is the number on Filippo's t-shirt?} & {\textbf{Text}\newline
    Luca tosses two fair six-sided dice.\newline
    \textbf{Question} \newline
    Finish the sentence adding one of the following options
    - higher than
    - smaller than
    - equal to
    "The probability that the sum of the two dice is 12 is ... probability that the sum is 2."
    }\\
 \end{supertabular}
 \end{center}

\clearpage

\section{Model Naming Summary}
\label{app:model_naming}
\begin{center}
    \tablefirsthead{%
		\toprule
		   Identifier   &  Huggingface Name & N. Params & (pre-trainig) Language & (fine-tuning) Language \\
		\midrule}
	\tabletail{%
		\bottomrule
	}
	\tablelasttail{\bottomrule}
 \tablecaption{Model identifiers along with their Huggingface model name, number of parameters and the most occuring language in pre-training and fine-tuning data.}
        \begin{supertabular}{m{4cm}m{5cm}m{1cm}m{2cm}m{2cm}}

    \mixtral{}  & mistralai/Mixtral-8x7B-Instruct-v0.1 & 47 B & multilingual & multilingual \\
    \mistral{} & mistralai/Mistral-7B-Instruct-v0.2 & 7 B & multilingual & multilingual \\
    \midrule
    \textit{llama 2 7b chat} & meta-llama/Llama-2-7b-chat-hf & 7 B & English & English \\
    \llamadue{} & meta-llama/Llama-2-70b-chat-hf & 70 B & English & English \\
    \llamantino{} & swap-uniba/LLaMAntino-2-70b-hf-UltraChat-ITA & 70 B & English & Italian \\
    \camoscio{} & ai4text/camoscio-70-b & 70 B & English & Italian \\
    \midrule
    \llamatreotto{} & meta-llama/Meta-Llama-3-8B-Instruct & 8 B & English & English \\
    \llamatresett{} & meta-llama/Meta-Llama-3-70B-Instruct & 70 B & English & English \\
    \midrule
    \llamatreunootto{} & meta-llama/Meta-Llama-3.1-8B-Instruct & 8 B & English & English \\
    \llamatreunosett{} & meta-llama/Meta-Llama-3.1-70B-Instruct & 70 B & English & English \\
    \llamatreunoquattro{} & meta-llama/Meta-Llama-3.1-405B-Instruct & 70 B & English & English \\
    \midrule
    \anita{} & swap-uniba/LLaMAntino-3-ANITA-8B-Inst-DPO-ITA & 8 B & English & Italian \\
    \minerva{} & sapienzanlp/Minerva-3B-base-v1.0 & 3 B & Italian & N/A \\ 
\end{supertabular}
\end{center}

\clearpage

\section{Invalsi Mate Human Evaluation}
\label{app:human_eval}
\begin{figure}[!t]
    \centering
    \includegraphics[width=\linewidth]{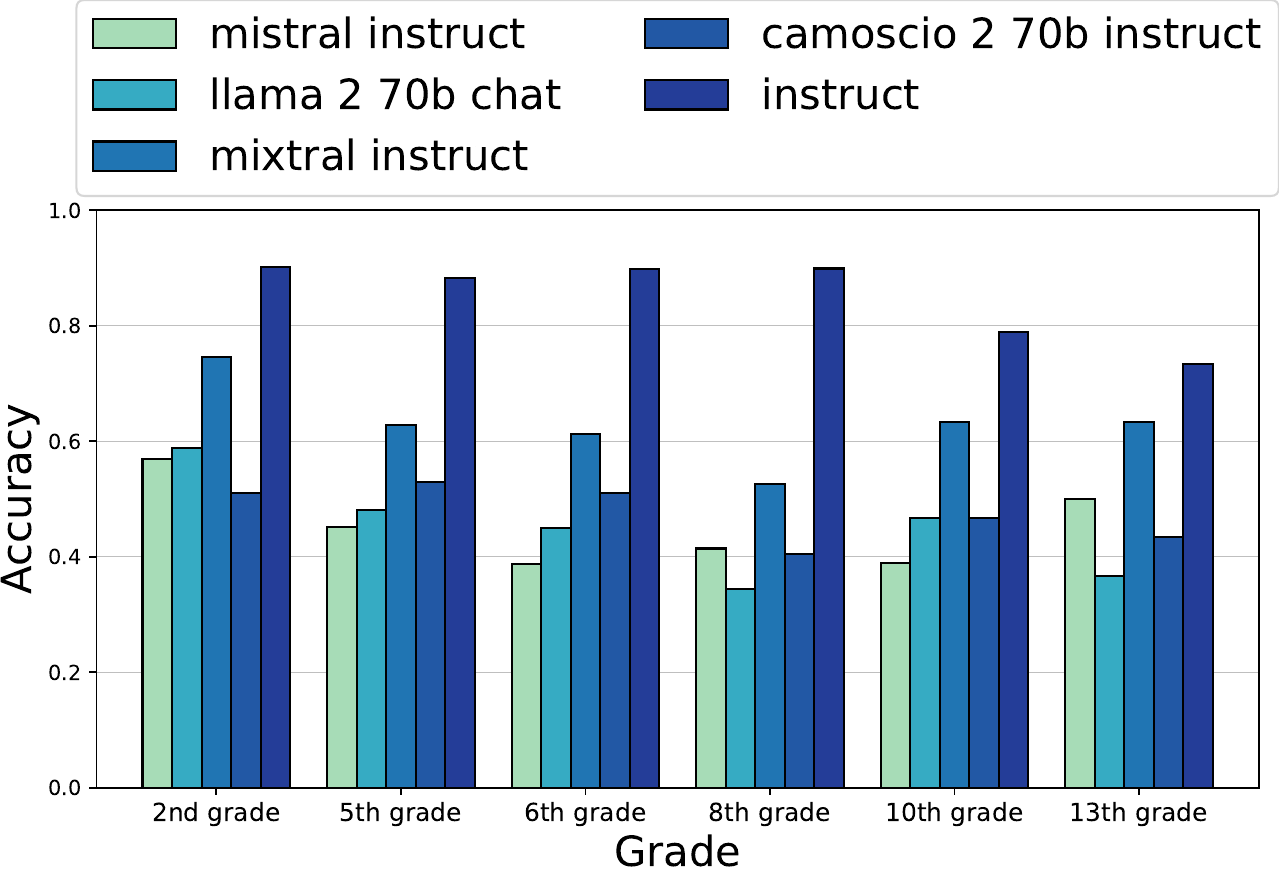}
    \caption{Performance of different Language Models on Invalsi MATE per grade level humanly assessed.}
    \label{fig:invalsi_mate_manual_perf_grade_dist}
\end{figure}

\begin{table*}[ht]
    \centering
\begin{tabular}{lccccc}
\toprule
Question Type & \textit{ALL} & \ftg{} & \mc{} & \num{} & \tf{} \\
N. Questions & 420 & 20 & 244 & 54 & 102 \\
\midrule
Model & \multicolumn{5}{c}{Accuracy} \\
 
\midrule
llama 3 70b instruct & 0.86 & 0.63 & 0.86 & 0.83 & 0.98 \\ 
mixtral instruct        & 0.62 & 0.37 & 0.63 & 0.63 & 0.65 \\
camoscio 2 70b instruct & 0.48 & 0.32 & 0.47 & 0.49 & 0.52 \\
llama 2 70b chat        & 0.45 & 0.37 & 0.43 & 0.44 & 0.56 \\
mistral instruct v0.2   & 0.44 & 0.21 & 0.44 & 0.42 & 0.56 \\
\bottomrule
\end{tabular}
    \caption{Models 0-Shot accuracy on Invalsi MATE, human evaluation.}
    \label{tab:invalsi_mate_human}
\end{table*}

Following likelihood based evaluation, we are interested in knowing the ability of Language models in answering questions by properly generating a complete answer, this is mostly interesting for \MATE{} since these questions require explicit reasoning and moreover, the evaluation on \ITA{} is too time consuming due to the long text passages providing the contexts to the questions, therefore we limit this analysis to \MATE{}.

For this open-ended setting we evaluate a subset of the language models tested so far: \mistral{}, \mixtral{}, \llamadue{}, \camoscio{}, and \llamatresett{}, we devise a distinct prompt for each question type and prompt all the tested language models to assess how well they understand Mathematical reasoning and Language in Italian.

We perform the evaluation in 0-shot fashion, i.e. each model is asked to answer the question and we use the chain of thought approach \citep{stepbystep} in that we prefix the language models answer by adding the words ``Ragioniamo passo passo'' (literally: ``Let's reason step by step'') to have it output explicit reasoning steps before providing an answer.

We remark that we don't use a chat based approach where we inject further requests one after the other, because some of the models we test are not meant for chat but are only instruction fine-tuned.

\Cref{tab:invalsi_mate_human} shows the accuracy achieved by LLMs we manually inspected. Models' ranking is kept and we observe that models fine-tuned in Italian have a lesser ability to clearly frame the answer, i.e. ``the correct answer is A''\footnote{In Italian, ``la risposta esatta è A''}, with respect to those fine-tuned in English or multilingual datasets, even though the correct answer is returned.

We argue that the main reason for this issue is the content and more importantly the size of the Italian fine-tuning datasets, which are smaller and of lower quality.

We also look at the performance distribution by grade, reported in \Cref{fig:invalsi_mate_manual_perf_grade_dist} which shows similar trends to what observed for likelihood based evaluations.

\clearpage

\section{Pattern Matching Evaluation}
\label{app:pattern_match_eval}
\begin{table*}[ht]
\centering
\begin{tabular}{lrrrrr}
\toprule
 & All & sentence completion & multiple choice & number & true/false \\
model &  &  &  &  &  \\
\midrule
mixtral instruct        & 0.53 & 0.00 & 0.60 & 0.42 & 0.57 \\
llama 2 70b chat        & 0.42 & 0.00 & 0.46 & 0.38 & 0.48 \\
mistral instruct v0.2   & 0.42 & 0.00 & 0.46 & 0.35 & 0.52 \\
llama 2 13b chat        & 0.24 & 0.00 & 0.27 & 0.12 & 0.46 \\
llama 2 7b chat         & 0.22 & 0.00 & 0.27 & 0.20 & 0.48 \\
camoscio 2 70b instruct & 0.13 & 0.00 & 0.14 & 0.96 & 0.39 \\
\bottomrule
\end{tabular}
\caption{Models 0-Shot accuracy on Invalsi MATE, pattern matching-based evaluation.}
\label{tab:invalsi_mate}
\end{table*}

\Cref{tab:invalsi_mate} shows the models accuracy on the \MATE{} benchmark, both the global accuracy when measuring all the question types together, as well as the accuracy on each separate question type. 
We notice how performance is heavily dependent on the type of questions, indeed the models perform very poorly on "Fill the gap" tasks, as shown by the "completa frase" columns and on True - False tasks they are also mostly close to random performance when not worse.
On the contrary, they show strong performance on both multiple choice and number answers where the strongest model, \emph{mixtral}, achieves up to 61.76\% accuracy. 
This evaluation is done by extracting the answers automatically (i.e., matching answer-template patterns with the text).

However, seeing the weak performance on True - False questions and after a manual inspection of the sentences we performed a more accurate human-made evaluation.

\clearpage

\section{Prompt Details}
\label{app:prompts}

\begin{figure*}[t]
    \centering
    \begin{subfigure}{.4\textwidth}
    \codesnippet{
        \footnotesize
        [INST] <<SYS>>

        Write in Italian as a very deductive and precise student.

        <</SYS>>

        Dato un testo (TESTO) completalo come richiesto e dai una risposta finale. [/INST]

        TESTO:

        \{testo\}

        DOMANDA:

        \{domanda\}

        Ragioniamo passo passo"
        }
        \caption{Fill the Gap}
        \label{subfig:math_llama_chat_fill_gap}
    \end{subfigure}\hspace{0.5cm}%
    \begin{subfigure}{.4\textwidth}
        \codesnippet{
            \footnotesize
        [INST] <<SYS>>

        Write in Italian as a very deductive and precise student.

        <</SYS>>

        Dato un testo (TESTO) scegli la risposta alla domanda (DOMANDA) seguente tra quelle disponibili e dai una risposta finale. [/INST]

        TESTO:

        \{testo\}

        DOMANDA:

        \{domanda\}

        Ragioniamo passo passo}
        \caption{Multiple Choice}
        \label{subfig:math_llama_chat_multiple}
    \end{subfigure}\vspace{0.2cm}
    \begin{subfigure}{.4\textwidth}
        \codesnippet{
        \footnotesize
        [INST] <<SYS>>

        Write in Italian as a very deductive and precise student.

        <</SYS>>

        Dato un testo (TESTO) calcola il numero come richiesto nella domanda (DOMANDA) all fine dai come risposta finale solo un numero. [/INST]

        TESTO:

        \{testo\}

        DOMANDA:

        \{domanda\}

        Ragioniamo passo passo}
        \caption{Number}
        \label{subfig:math_llama_chat_number}
    \end{subfigure}\hspace{0.5cm}%
    \begin{subfigure}{.4\textwidth}
        \codesnippet{
        \footnotesize
        [INST] <<SYS>>

        Write in Italian as a very deductive and precise student.

        <</SYS>>

        Dato un testo (TESTO) indica se la frase (FRASE) che lo segue è vera o falsa e dai una risposta finale. [/INST]

        TESTO:

        \{testo\}

        FRASE:

        \{domanda\}

        Ragioniamo passo passo}
        \caption{True/False}
        \label{subfig:math_llama_chat_truefalse}
    \end{subfigure}

    \caption{Prompt templatess for the Invalsi MATH dataset for the LLama Chat and LLaMantino models. 
    The placeholders ``[testo]'' and ``[domanda]'' are replaced with the two pieces of text that respectively give information about the problem and then ask the question.
    In (\subref{subfig:math_llama_chat_fill_gap}) for the fill the gap questions, in (\subref{subfig:math_llama_chat_multiple}) for the multiple choice questions, in (\subref{subfig:math_llama_chat_number}) for number questions and in (\subref{subfig:math_llama_chat_truefalse}) for True/False questions.}
    \label{fig:math_llama_chat_prompts}
    \end{figure*}

\begin{figure*}[t]
    \centering
    \begin{subfigure}{.4\textwidth}
    \codesnippet[3.5cm]{
        \footnotesize
        [INST] Dato un testo (TESTO) completalo come richiesto e dai una risposta finale. [/INST]

        TESTO:

        \{testo\}

        DOMANDA:

        \{domanda\}

        Ragioniamo passo passo}
        \caption{Fill the Gap}
        \label{subfig:math_mistral_fill_gap}
    \end{subfigure}\hspace{0.5cm}%
    \begin{subfigure}{.4\textwidth}
        \codesnippet[3.5cm]{
            \footnotesize
            [INST] Dato un testo (TESTO) scegli la risposta alla domanda (DOMANDA) seguente tra quelle disponibili e dai una risposta finale. [/INST]

            TESTO:

            \{testo\}

            DOMANDA:

            \{domanda\}

            Ragioniamo passo passo}
        \caption{Multiple Choice}
        \label{subfig:math_mistral_multiple}
    \end{subfigure}\vspace{0.2cm}
    \begin{subfigure}{.4\textwidth}
        \codesnippet[3.5cm]{
        \footnotesize
        [INST] Dato un testo (TESTO) calcola il numero come richiesto nella domanda (DOMANDA) all fine dai come risposta finale solo il numero. [/INST]

        TESTO:

        \{testo\}

        DOMANDA:

        \{domanda\}

        Ragioniamo passo passo}
        \caption{Number}
        \label{subfig:math_mistral_number}
    \end{subfigure}\hspace{0.5cm}%
    \begin{subfigure}{.4\textwidth}
        \codesnippet[3.5cm]{
        \footnotesize
        [INST] Dato un testo (TESTO) indica se la frase (FRASE) che lo segue è vera o falsa e dai una risposta finale. [/INST]

        TESTO:

        \{testo\}

        FRASE:

        \{domanda\}

        Ragioniamo passo passo}
        \caption{True/False}
        \label{subfig:math_mistral_truefalse}
    \end{subfigure}
    \caption{Prompts for the Invalsi MATH dataset for Mistral and Mixtral and DanteLLM models. In (\subref{subfig:math_mistral_fill_gap}) for the fill the gap questions, in (\subref{subfig:math_mistral_multiple}) for the multiple choice questions, in (\subref{subfig:math_mistral_number}) for number and in (\subref{subfig:math_mistral_truefalse}) for True/False questions.}
    \label{fig:math_mistral_prompts}
\end{figure*}

\end{document}